%% file: main.tex
\documentclass[10pt,twocolumn,letterpaper]{article}

\usepackage{iccv}
\usepackage{times}
\usepackage{epsfig}
\usepackage{graphicx}
\usepackage{amsmath}
\usepackage{amssymb}

\usepackage{mathtools}
\usepackage{amsmath,amssymb,amsfonts}
\usepackage{textcomp}
\usepackage{picinpar}
\usepackage{url}
\usepackage{flushend}
\usepackage{colortbl}
\usepackage{soul}
\usepackage{multirow}
\usepackage{pifont}
\usepackage{color}
\usepackage{alltt}
\usepackage{enumerate}
\usepackage{siunitx}
\usepackage{epstopdf}
\usepackage{pbox}
\usepackage{caption}
\usepackage{subcaption} 
\usepackage{stmaryrd}
\usepackage{accents}
\usepackage{booktabs}
\usepackage{algorithm} 
\usepackage{algpseudocode} 
\usepackage{esvect}
\usepackage{bm}
\usepackage{mwe}
\usepackage{amsfonts} 
\usepackage{float}
\usepackage{bbold}
\usepackage{gensymb}

\algrenewcommand\algorithmicrequire{\textbf{Input:}}
\algrenewcommand\algorithmicensure{\textbf{Output:}}

\newcounter{phase}[algorithm]
\newlength{\phaserulewidth}
\newcommand{\setphaserulewidth}{\setlength{\phaserulewidth}}
\newcommand{\phase}[1]{%
  \vspace{-1.25ex}
  \Statex\leavevmode\llap{\rule{\dimexpr\labelwidth+\labelsep}{\phaserulewidth}}\rule{\linewidth}{\phaserulewidth}
  \Statex\strut\refstepcounter{phase}\textit{Phase~\thephase~--~#1}
  \vspace{-1.25ex}\Statex\leavevmode\llap{\rule{\dimexpr\labelwidth+\labelsep}{\phaserulewidth}}\rule{\linewidth}{\phaserulewidth}}
\setphaserulewidth{.7pt}

\makeatletter 
\newcommand{\vE}{\vec{\pmb{E}}\@ifnextchar{^}{\,}{}}
\makeatother 

\makeatletter
\newcommand{\thickhline}{
    \noalign {\ifnum 0=`}\fi \hrule height 1pt
    \futurelet \reserved@a \@xhline
}
\makeatother

\newcommand{\comment}[1]{}
\newcommand{\cmark}{\ding{51}}%
\newcommand{\xmark}{\ding{55}}%

\comment{
\usepackage[utf8]{inputenc}
\usepackage{fourier} 
\usepackage{array}
\usepackage{makecell}

}

\usepackage[pagebackref=true,breaklinks=true,letterpaper=true,colorlinks,bookmarks=false]{hyperref}

 \iccvfinalcopy 

\def\iccvPaperID{10679} 

\ificcvfinal\fi

\begin{document}

\title{Balanced Supervised Contrastive Learning for \\
Few-Shot Class-Incremental Learning}

\author{In-Ug Yoon\\
KAIST\\
{\tt\small iuyoon@rit.kaist.ac.kr}
\and
Tae-Min Choi\\
KAIST\\
{\tt\small tmchoi@rit.kaist.ac.kr}
\and
Young-Min Kim\\
Samsung Research\\
{\tt\small ym1012.kim@samsung.com}
\and
Jong-Hwan Kim\\
KAIST\\
{\tt\small jhkim@rit.kaist.ac.kr}
}

\maketitle
\ificcvfinal\thispagestyle{empty}\fi

\begin{abstract}
Few-shot class-incremental learning (FSCIL) presents the primary challenge of balancing underfitting to a new session's task and forgetting the tasks from previous sessions. To address this challenge, we develop a simple yet powerful learning scheme that integrates effective methods for each core component of the FSCIL network, including the feature extractor, base session classifiers, and incremental session classifiers. In feature extractor training, our goal is to obtain balanced generic representations that benefit both current viewable and unseen or past classes. To achieve this, we propose a balanced supervised contrastive loss that effectively balances these two objectives. 
In terms of classifiers, we analyze and emphasize the importance of unifying initialization methods for both the base and incremental session classifiers. 
Our method demonstrates outstanding ability for new task learning and preventing forgetting on CUB200, CIFAR100, and miniImagenet datasets, with significant improvements over previous state-of-the-art methods across diverse metrics. We conduct experiments to analyze the significance and rationale behind our approach and visualize the effectiveness of our representations on new tasks. Furthermore, we conduct diverse ablation studies to analyze the effects of each module.
\end{abstract}

\input{1_introduction.tex}

\input{2_related_works.tex}
\input{3_problem_settings.tex}

\input{4_methods.tex}
\input{5_experiments.tex}

\input{6_conclusion.tex}

{\small
\bibliographystyle{ieee_fullname}
\bibliography{egbib}
}

\clearpage
\appendix
\input{supp}

\end{document}

%% file: 1_introduction.tex
\section{Introduction}
\label{sec:intro}

\begin{figure*}[t]
    \centering 
    \begin{subfigure}{6.5cm}
    \includegraphics[width=6.5cm]{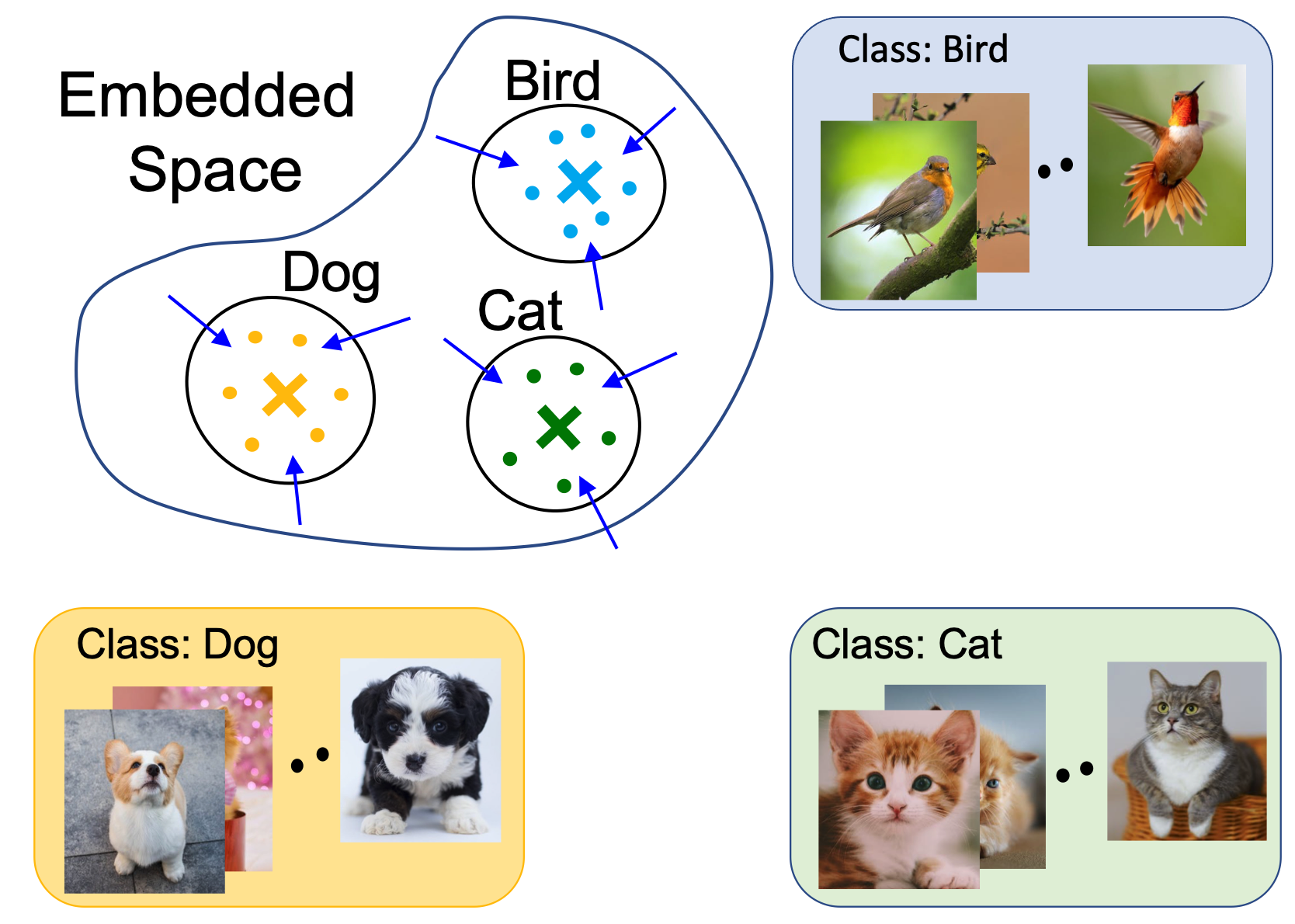}
    \caption{Directions of class-wise losses}
    \label{fig:loss_direction_a}
    \end{subfigure}
    \quad
    \begin{subfigure}{8.2cm}
    \includegraphics[width=8.2cm]{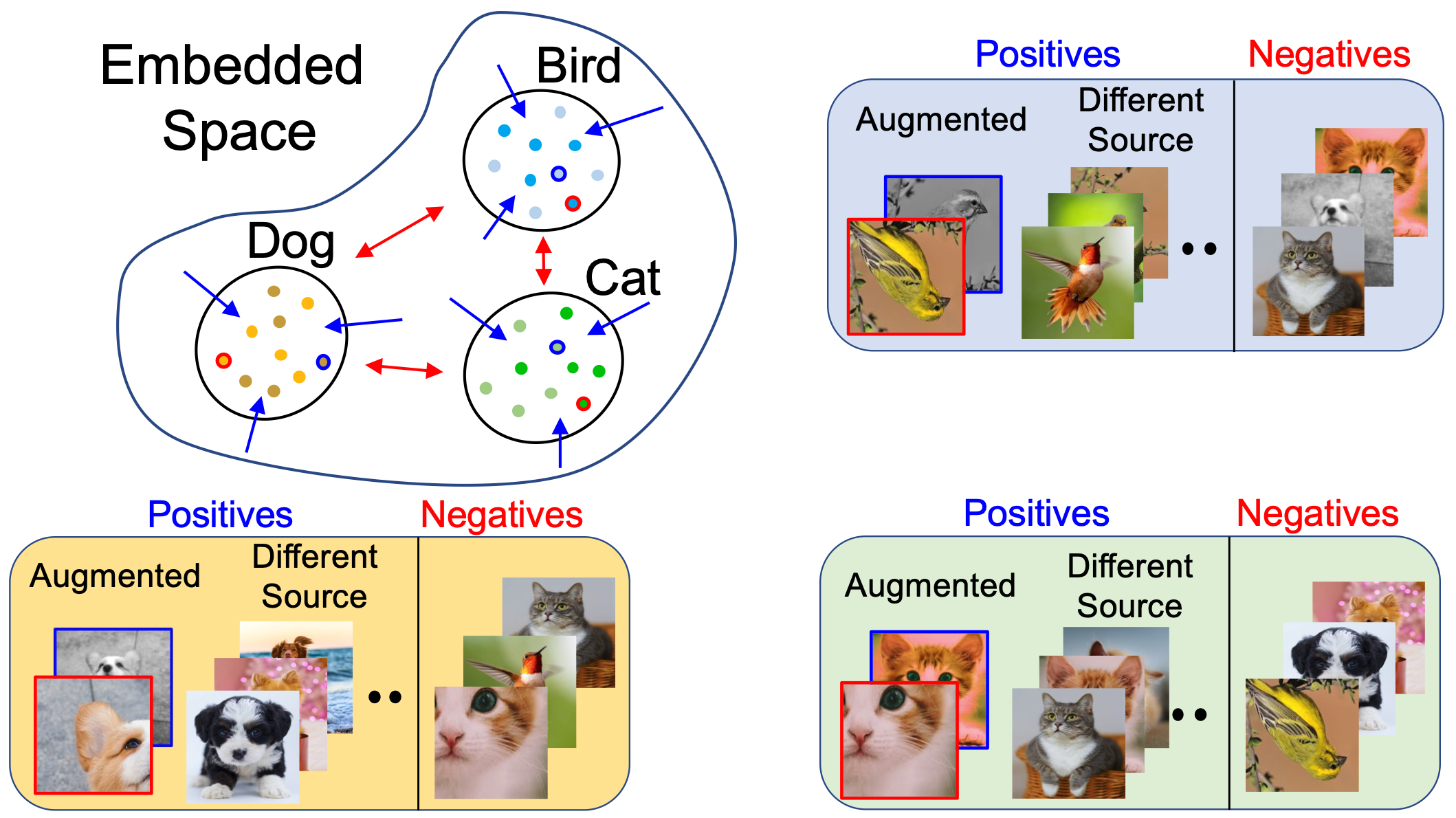}
    \caption{Directions of contrastive losses}
    \label{fig:loss_direction_b}
    \end{subfigure}
    \caption{
    Schematics explaining the objectives of the losses that are used during the feature extractor training.
    (a) Class-wise losses: Each class has a classifier, so the features of each class are aggregated to the corresponding classifier (x marks). 
    (b) Contrastive losses: 
    Multi-viewed batch is composed by adding augmented images (having features with slightly different colors) to the original batch.
    Each box describes the composition of multi-viewed batch when the anchor image (red borderline) is chosen (anchor feature is the red-borderline point within the region corresponding to anchor label, and blue borderline denotes the augmented pair of the anchor). 
    Then, positives are aggregated to each other while negatives are dispersed away.
    }
    \label{fig:loss_direction}
\end{figure*}

Incremental learning with limited data is a common scenario in real-world applications. Setting up deep learning algorithms for image classification tasks, in particular, can be challenging, especially when new classes are continuously being introduced with only a few samples available for training.  Few-shot class-incremental learning (FSCIL) aims to address this challenge \cite{cheraghian2021semantic,dong2021few,chen2020incremental}.
FSCIL involves sequentially given sessions of tasks. The base (first) session has a large training set, while individual incremental sessions have a limited amount of data. The test set of each session contains all the classes seen so far.
The primary issue with FSCIL is the trade-off between learning new knowledge and preventing forgetting of the previous sessions.
Various approaches have been proposed to address the challenge of few-shot class-incremental learning. These methods include knowledge distillation \cite{dong2021few,cheraghian2021semantic}, coreset-based replay \cite{liu2022few}, and parameter regularization to prevent forgetting \cite{chen2020incremental}.
Another line of work involves training a feature extractor with meta-learning by composing virtual training episodes with a base session training set \cite{hersche2022constrained,zhou2022few}.
Methods such as re-balancing the previous and new classifier weights using the attention model \cite{zhang2021few}, giving a margin between the classifiers based on model update speed difference \cite{zou2022margin}, and projecting classifiers to subspace for richer representations have been presented \cite{cheraghian2021synthesized}.


However, current state-of-the-art FSCIL methods have shown significant decrement in accuracy during sessions and low accuracy being measured only on incremental classes.
To enhance overall performance, we propose a novel learning scheme that takes into account both the feature extractor training method for producing \textit{generic representations} (which we refer to representations being effective not only for the current viewable classes but also for unseen or past ones) and the initialization method for classifiers in the base and incremental sessions.
Regarding the feature extractor, we first identify limitations in previous works. These approaches typically train feature extractors based on class-wise losses, such as cross-entropy (CE) loss, as shown in Fig. \ref{fig:loss_direction_a}. The CE loss aggregates each class's features into the corresponding classifier, resulting in a feature extractor that achieves high classification performance for current labeled classes.
However, in previous few-shot class-incremental learning (FSCIL) methods, the representations extracted from such feature extractors exhibited insufficient performance on unseen or past classes during the incremental sessions \cite{zhang2021few, tao2020few, cheraghian2021semantic, zhou2022forward}.
Thus, our aim is to design a training scheme that enables the feature extractor to extract visual representations having the characteristics of generic representations.
In contrast to using the class-wise approach, we design a feature extractor pre-training scheme that aims to generate generic representations using the advantages of contrastive losses, as illustrated in Fig. \ref{fig:loss_direction_b}.  

The most representative contrastive loss for the supervised task is the supervised contrastive (SupCon) loss \cite{khosla2020supervised}. SupCon first creates a multi-viewed batch that includes additional images augmented from the images in the batch, with each image having a single additional augmented image. Then, SupCon gathers the features of images from the same classes. While the generalization ability of representations is originally derived from gathering the features of differently augmented images from the same source image \cite{chen2020simple,feng2019self,chen2020improved,doersch2015unsupervised}, augmented images and images from the same class with different source images are weighted equally in SupCon. Moreover, the number of augmented images within the multi-view batch is relatively small. 
To this end, we claim that increasing the proportion of augmented images during representation learning may enhance generalization ability. However, increasing the proportion of augmented images results in a reduction in the proportion of images from the same class, which play a role in enhancing the model's ability to classify that class. Therefore, we propose the balanced supervised contrastive (BSC) loss to control the proportion of augmented images used for representation learning, thereby balancing the effectiveness of the extracted representations for both viewable and unviewable classes.
We emphasize that balancing the generalization abilities of representations is especially critical for FSCIL because the trade-off between learning new knowledge and preventing forgetting about the past is a major concern.


Since the pre-training stage is focused on enhancing the generalization of representations rather than acquiring knowledge for the base session itself, we add a fine-tuning stage to update the feature extractor and base classifiers to acquire domain-specific knowledge.
An important consideration to note is that, similar to the incremental session classifiers, we initialize the base classifiers as the average of features corresponding to each class. This is to ensure consistency between the base and incremental classifiers, as analyzed and explained in Section \ref{sec:fine_tune}.

We achieve state-of-the-art results on all three commonly used datasets, using diverse metrics. 
Note that we propose two additional metrics, new task learning ability (NLA) and base task maintaining ability (BMA), to address the limitations of the commonly used FSCIL metric that only measures the performance drop during the incremental sessions without precise observations on the performance changes of both base and incremental session classes in class-incremental settings.
Furthermore, we conduct ablation studies to analyze the importance of each module used in the proposed framework. In short, our contributions are:
\begin{itemize}
\item{We propose the BSC loss for training the feature extractor to obtain balanced generic representations, which are vital for FSCIL.}
\item{We analyze the importance of unifying the initialization methods for base and incremental classes.}
\item{Our approach is validated through experiments on diverse datasets and metrics, including our proposed metrics NLA and BMA, demonstrating its effectiveness in learning new tasks and preventing forgetting. The results show a significant improvement compared to previous state-of-the-art methods.}
\end{itemize}





%% file: 2_related_works.tex
\section{Related Work}
\label{sec:related_work}

\textbf{Few-shot class-incremental learning.} 
FSCIL limits the amount of training data available for each incremental session in the CIL setting, making it useful for new tasks with limited data availability. The main issue with FSCIL is the trade-off between learning new knowledge for the new task and preventing forgetting of past knowledge. The mainstream of research attempts to develop methods to mitigate forgetting.
In \cite{tao2020few}, the concept of neural gas was presented to preserve the topology of features between the base and new session classes. Knowledge distillation has also been utilized to prevent catastrophic forgetting by using the previous model as a teacher model \cite{dong2021few, cheraghian2021semantic}.
Data replay involves saving a certain amount of previous task data as an exemplar set and replaying it during incremental sessions \cite{liu2022few}. Additionally, expanding networks for each knowledge were introduced in \cite{ji2022memorizing}.

Other lines of work focus on feature extractor training methods for incremental settings. Loss terms based on L2 distance have been designed to train the network in \cite{chen2020incremental}.
Creating episodic scenarios with base session data and training the feature extractor with meta-learning-based methods were attempted to increase adaptation ability to new tasks \cite{zhou2022few, hersche2022constrained}.
Preserving areas in the embedding space for upcoming classes has also shown performance enhancement \cite{zhou2022forward}. Some works freeze the feature extractor in incremental sessions and balance the classifiers between past and new classes, using attention models to balance the incrementally drawn classifiers in \cite{zhang2021few}.
However, the above approaches have yet to attempt to obtain general visual representations that enable the extraction of meaningful representations for both seen and unseen classes. Self-supervised learning has also been attempted for FSCIL but using a feature extractor pre-trained on an external dataset \cite{ahmad2022variable}.


\textbf{Self-supervised learning and supervised contrastive learning}
Recently, self-supervised learning has shown significant achievements in learning effective visual representations \cite{doersch2015unsupervised,dosovitskiy2014discriminative,chen2020improved,feng2019self,chen2020simple}. Research has shown that considering the contrastive relationship between the features of images leads to good representation learning, even without using label information \cite{chen2020simple,chen2020improved}. 
Furthermore, supervised contrastive (SupCon) learning provides a method to leverage the label information on top of contrastive self-supervised learning \cite{khosla2020supervised}. 
In addition to SupCon, there are other approaches such as parametric contrastive learning (PaCo) \cite{cui2022generalized}, which incorporates parametric class-wise learnable centers, and THANOS \cite{chen2022perfectly}, which includes repulsion between the feature of the anchor's augmented image and the feature of other positives within the multi-viewed batch.
However, the representations trained with these methods exhibit limited generality for new unseen classes, as discussed in our experiments section.


\comment{
\textbf{Regularization methods to prevent overfitting}
Various regularization methods have been proposed to address overfitting, such as drop-out\cite{srivastava2014dropout}, batch normalization\cite{ioffe2015batch}, L1/L2 regularization\cite{nowlan2018simplifying}, and data augmentation\cite{cubuk2018autoaugment}.
The regularizing predictive distribution also effectively reduces overfitting since it contains most core knowledge of models. 
Recently class                   -wise self-knowledge distillation (cs-kd) has shown effectiveness in preventing overconfident predictions and reducing intra-class variation by matching the predictive distribution of the network between different samples of the same label\cite{yun2020regularizing}.
}

%% file: 3_problem_settings.tex
\section{Problem Set-up}
\label{sec:problem_setup_preliminaries}

\label{sec:problem_formulation}
FSCIL is given with sequential tasks $\mathcal{D}=\{\mathcal{D}^1,\mathcal{D}^2,...,\mathcal{D}^T\}$, the dataset for session \textit{t} is defined as $\mathcal{D}^t=\{\mathcal{D}^t_{tr},\mathcal{D}^t_{te}\}$.
Train set of session \textit{t}, $D^t_{tr}$ consists of class labels $\mathcal{C}^t_{tr}=\{c^t_1,...,c^t_{n^t}\}$, where $n^t$ is the number of classes in session $t$.
Note that different sessions have no overlapped classes, $\textit{i.e.}$ $\forall i,j$ and $i\neq j$, $\mathcal{C}^i\cap\mathcal{C}^j=\varnothing$.
Test set of session \textit{t}, $D^t_{te}$ is composed of class labels, $\mathcal{C}^1\cup\mathcal{C}^2\cdot\cdot\cdot\cup\:\mathcal{C}^t$.
For convenience, we denote $\mathcal{C}^{i:j}=\mathcal{C}^i\cup\cdot\cdot\cdot\cup\:\mathcal{C}^j$ for $i<j$.
For $\textit{N}$-way $\textit{K}$-shot FSCIL setting, $|\mathcal{D}^t_{tr}|=K$, $n^t=N$ for $t>1$. 
For example, in the popular benchmark dataset CUB200, there are 100 classes in the base session and 100 classes for incremental sessions. For 10-way 5-shot setting, the number of sessions $T=11$, $K=5$ and $|\mathcal{C}^t|=10$ for $t>1$.

Feature extractor is expressed as $f_{\theta}(\cdot)$, where $\theta$ represents network parameters. 
The input is encoded to $\mathbb{R}^D$embedded space by passing the feature extractor. Each class $\textit{c}$ has a classifier $w_c\in\mathbb{R}^D$ for the classification.
We denote $w_{\mathcal{C}}=\{w_i$  for $i\in\mathcal{C}\}$ as the classifiers of individual classes.

\begin{figure*}[t]
    \centering 
    \includegraphics[width=13.0cm]{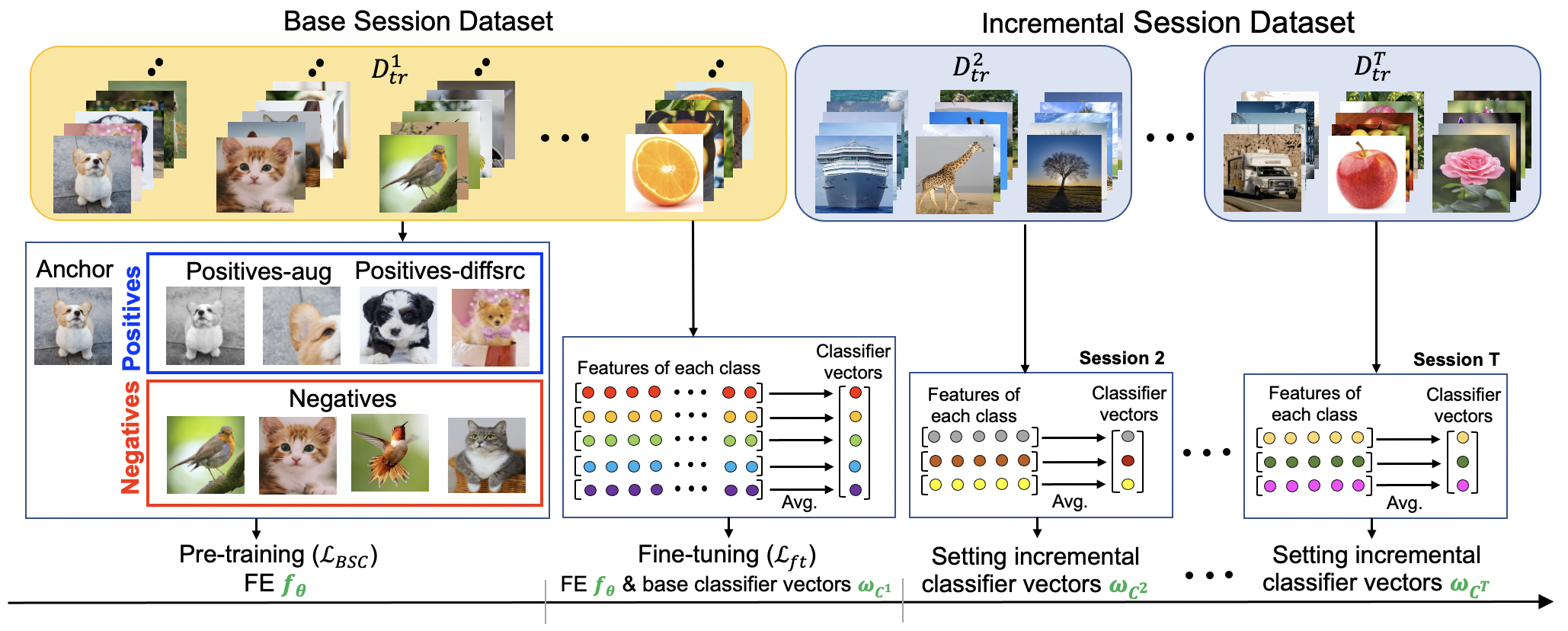}
    \caption{
    Overall schematics of the proposed FSCIL method. 
    First, the feature extractor is pre-trained to achieve generic representations using the massive base training set $D_{tr}^1$.
    Augmented or having the same label with anchor image are set as positive and else are set to be negative within the batch.
    Then the model is updated based on the contrastive loss.
    Second, we fine-tune the model using $D_{tr}^1$ to learn the specific domain knowledge of the dataset. The base classifiers $w_{\mathcal{C}^1}$ are first initialized as the feature means of the corresponding class before the fine-tuning process. In the incremental sessions, we simply set the classifiers $w_{\mathcal{C}^t}$ as the mean feature of the corresponding classes, without any further updates.
    }
    \label{fig:schematics}
\end{figure*}

%% file: 4_methods.tex
\section{Method}
\label{sec:method}

In this section, we introduce the overall training scheme of our proposed method for FSCIL as shown in Fig. \ref{fig:schematics}, also described in Algorithm (Appendix \ref{app:sec:alogrithm}). We first describe the feature extractor pre-training process in Section \ref{sec:pre_train}. 
Then, we introduce fine-tuning procedure in Section \ref{sec:fine_tune}, and finally, the definition of classifiers for incremental sessions is described in Section \ref{sec:inc_sessions}.

\subsection{Pre-training feature extractor for generic representations}
\label{sec:pre_train}

Our primary goal is to train a feature extractor capable of producing visual representations that can extract meaningful features not only from current classes, but also from past and unseen classes in all sessions. These representations are commonly referred to as \textit{generic representations}. The majority of current research focuses on training feature extractors using a class-wise loss, which may perform well on trained classes but not designed to learn generic representations. Recently, supervised contrastive (SupCon) learning \cite{khosla2020supervised} has proven to be effective in training feature extractors to achieve generic representations. The transferability of these representations has been tested on various tasks, including classification, segmentation, and object detection.

Originally, the enhancement of generalization abilities within the representations was based on gathering only the features of augmented images \cite{chen2020simple,feng2019self,chen2020improved,doersch2015unsupervised}. However, in SupCon learning, equal weights are assigned to augmented images and same-class images. Moreover, the number of augmented images in the multi-viewed batch is comparatively limited, thereby minimizing their impact on the overall procedure. To emphasize the effect of augmented images, we propose balanced supervised contrastive (BSC) loss. For each image within the multi-viewed batch (which we simply call the \textit{anchor}), previous approaches denote the images with the same class as the \textit{anchor} as \textit{positives}, and the rest as \textit{negatives}. In BSC, we further split the \textit{positives} into \textit{positives-aug}, which have the same source image for the augmentation as the \textit{anchor}, and \textit{positives-diffsrc}, which have a different source image for the augmentation. Then, during representation learning, the BSC loss emphasizes the effects of \textit{positives-aug} more by utilizing balancing scale factors and multiple augmented images.


For the formulations, we first describe the multi-viewed batch.
Given training batch data $\mathcal{D}_{tr}=\{x_n,y_n\}_{n=1}^N$, we compose the multi-viewed batch $J_m\equiv \{x_j,y_j\}_{j=1}^{mN}$ as
\begin{equation}
\begin{split}
\resizebox{1.0\hsize}{!}{$ 
    x_j= 
\begin{dcases}
    Aug^{(1)}(x_j),& \text{if } j\in\{1,..,N\}\\
    ... \\
    Aug^{(m)}(x_{j-(m-1)N}),              & \text{if } j\in\{(m-1)N+1,..,mN\}\\
\end{dcases}
$}
\end{split}
\label{eq:multiview_twotype}
\end{equation}
where $m$ is the number of multiple augmentations for each source image and  $Aug^{(1)},...,Aug^{(m)}$ are augmentation modules which may differ from each other. 
For the \textit{anchor} $j\in\{1,...,|J_m|\}$ denoting the index for $J_m$, the major components within the multi-viewed batch are defined as
\begin{equation}
\begin{split}
\resizebox{.92\hsize}{!}{$
\begin{dcases}
    P(j)\equiv \{p\in A(j)| (j-p) \% m= 0 \}, & \textit{positives-aug} \\
    Q(j)\equiv \{q\in A(j)| y_q = y_j\} \setminus P(j), & \textit{positives-diffsrc}\\
    R(j)\equiv (A(j)\setminus (P(j) \cup Q(j)), & \textit{negatives}
\end{dcases}
$}
\end{split}
\label{eq:def_sets}
\end{equation}
 where $A(j)\equiv \{1,...,|J_m|\}\setminus\{j\}$ and $\%$ symbol denotes the remainder operator.
For the sake of convenience in our expressions, we define the softmax-based function as follows:
\begin{equation}
\begin{split}
l_{\theta,\phi}(x;j,J_m)=log\frac{exp(sim(h_\phi(f_\theta(x)),h_j)/\tau)}{\sum_{a\in A(j)}exp(sim(h_\phi(f_\theta(x)),h_j)/\tau)}.
\end{split}
\label{eq:softmax_based_fn}
\end{equation}
Here $h_j$ is defined as $h_j\equiv h_\phi(z_{j,\theta})=h_\phi(f_\theta(x_j))$, where
$z_{j,\theta}=f_\theta(x_j)$ is the encoded feature of input $x_j$ by feature extractor $f_\theta$ and
$h_\phi$ is a non-linear projection network that maps representations to the space where contrastive loss is applied.
For simplicity, we use the expression $h_j$ instead of $h_{j,\theta,\phi}$.
$\tau$ is the temperature parameter for scaling,
$sim(u,v)=u^\top v/\lVert{u}\rVert\lVert v\rVert$ denotes the dot product between $l_2$ normalized $u$ and $v$ (i.e., cosine similarity).

Finally, our proposed BSC loss for the feature extractor pre-training is expressed as follows:
\begin{equation}
\begin{split}
\resizebox{1.0\hsize}{!}{$\mathcal{L}_{BSC}(J_m;\theta,\phi)= \frac{1}{|J_m|}\sum_{j}\frac{-1}{\alpha |P(j)|+|Q(j)|}\qquad\qquad\qquad$}\\
\resizebox{0.9\hsize}{!}{$\times[\alpha\sum_{p\in P(j)}l_{\theta,\phi}(x_p;j,J_m)+\sum_{q\in Q(j)}l_{\theta,\phi}(x_q;j,J_m)]$}.
\end{split}
\label{eq:loss_pre}
\end{equation}
Note that we use the projection of features instead of the features themselves. 
By using such projection, the network is prevented from being biased towards the current dataset, thereby expanding the generality of features.

\subsection{Fine-tuning for the domain knowledge transfer}
\label{sec:fine_tune}
After pre-training, our focus shifts to training the model to acquire specific domain knowledge relevant to the dataset. To achieve this, we add a fine-tuning process since the feature extractor trained using Eq. \ref{eq:loss_pre} prioritizes generality over domain-specific knowledge. However, training networks with large amounts of base session training data can lead to poor generalization and overfitting. 
In order to further enhance the generalization of learned representations, we adopt the cs-kd loss \cite{yun2020regularizing} and modify it for our multi-viewed augmentation $J_m$ as follows:
\begin{equation}
\begin{split}
\resizebox{1.0\hsize}{!}{$\mathcal{L}_{cs}(J_m;\theta,w_{\mathcal{C}^1})= 
\frac{1}{mN} \sum_{j=1}^{mN}{KL(g({x_j}';\tilde{\theta}, \tilde{w}_{\mathcal{C}^{1}})||g(x_j;\theta, w_{\mathcal{C}^1}))}$},
\end{split}
\label{eq:cskdloss}
\end{equation}
where $\tilde{\theta}$ is a fixed copy of parameters, and KL denotes the KL divergence loss. Note that $g(\cdot)$ calculates the cross-entropy logit with a given feature extractor $f_\theta$ and classifiers $w_{\mathcal{C}}$.
${x_j}'$ is a randomly selected sample from $P(j)$, denoting the positive sample set of $x_j$.

Before updating the entire network, we initialize the classifiers for each base session class $c\in\mathcal{C}^1$ as the average vector of the corresponding class features, as follows:
\begin{equation}
\begin{split}
w_c = \frac{1}{|\mathcal{D}^1_{tr,c}|}\sum_{k\in \mathcal{D}^1_{tr,c}}{z_k/\lVert z_k \rVert},
\end{split}
\label{eq:base_clf_init}
\end{equation}
where $\mathcal{D}^1_{tr,c}\equiv \{k | (x_k,y_k)\in \mathcal{D}^1_{tr} \And y_k=c\}$.
Then the entire network is fine-tuned using the following loss:
\begin{equation}
\begin{split}
\mathcal{L}_{ft} = \mathcal{L}_{ce}(J_m;\theta,w_{\mathcal{C}^1}) + \lambda * \mathcal{L}_{cs}(J_m;\theta,w_{\mathcal{C}^1}),
\end{split}
\label{eq:loss_ft}
\end{equation}
where $\mathcal{L}_{ce}$ is the cross-entropy loss and $\lambda$ is the scaling hyperparameter.
Note that during the fine-tuning process, we do not use $\phi$ since the projection head is just to enhance the generality of feature extractor $f_\theta$ in Section \ref{sec:pre_train}.

The reason for not optimizing the base session classifiers from random initialization is as follows: when analyzing the features extracted from the optimized feature extractor using angular analysis, we found that the actual space occupied by features is narrow compared to the entire embedding space (This is explained in Appendix \ref{app:sec:angle_basetrain}.) However, classifiers that are optimized from random initialization are spread widely over the entire embedding space. 
Therefore, when using both random initialization and feature-mean initialization for classifiers, the majority of classified results will be concentrated on the mean-initialized classifiers, as we demonstrate through experimental results (Section \ref{sec:main_results}).
To address this, we initialize the classifiers of base sessions as the corresponding mean vectors before the fine-tuning process, since the classifiers for incremental classes should also be initialized as the corresponding mean feature vectors, as explained in Section \ref{sec:inc_sessions}.



\subsection{Setting classifiers for the incremental classes}
\label{sec:inc_sessions}
For the incremental sessions, we set the classifiers for the classes of the corresponding session as the average vector of the features of the training dataset, as follows:
\begin{equation}
\begin{split}
w_c = \frac{1}{|\mathcal{D}^t_{tr,c}|}\sum_{k\in \mathcal{D}^t_{tr,c}}{z_k / \lVert z_k \rVert},
\end{split}
\label{eq:inc_clf_init}
\end{equation}
where $c\in\mathcal{C}^t$ and $t$ is the session number.
The reason for using the mean feature vector is that a small training set size could easily cause overfitting if one optimizes the classifier from a random initialization. 
Note that this method does not require any further updates, which enables much faster adaptation to incremental sessions compared to methods that require updates in incremental sessions. Additionally, we demonstrate that utilizing pre-trained generic representations without further updates on incremental sessions leads to even better classification performance on the classes of incremental sessions than the methods that update the model on incremental sessions.

%% file: 5_experiments.tex
\section{Experiments}
\label{sec:experiments}
In this section, we quantitatively examine and analyze the performance of our proposed method.
Section \ref{sec:exp_setups} describes the experimental setups.
Then we show the main results of our approach in Section \ref{sec:main_results}.
To show the effects of each module utilized, we provide the ablation studies in Section \ref{sec:ablation_studies}.

\begin{table*}[!t]
\centering
\resizebox{0.86\textwidth}{!}{%
\begin{subtable}{1\textwidth}
\sisetup{table-format=4.0} 
    \renewcommand{\arraystretch}{1.1}
    \centering
    \begin{tabular}{l c c c c c c c c c l }
    \thickhline
    \multirow{2}{*}{\textbf{Method}} & \multicolumn{9}{c}{\textbf{Acc. in each session$\uparrow$ (\%) }} & \multirow{2}{*}{\textbf{PD($\downarrow$)}} \\ 
    \cline{2-10}
     & 1 & 2 & 3 & 4 & 5 & 6 & 7 & 8 & 9 \\
    \hline
     TOPIC\cite{tao2020few} & 61.31 & 50.09 & 45.17 & 41.16 & 37.48 & 35.52 & 32.19 & 29.46 & 24.42 & 36.89 \\
    CEC\cite{zhang2021few} & 72.00 & 66.83 & 62.97 & 59.43 & 56.70 & 53.73 & 51.19 & 49.24 & 47.63 & 24.37 \\
    LIMIT\cite{zhou2022few} & 72.32 & 68.47 & 64.30 & 60.78 & 57.95 & 55.07 & 52.70 & 50.72 & 49.19 & 23.13 \\
    FACT\cite{zhou2022forward} & 72.56 & 69.63 & 66.38 & 62.77 & 60.6 & 57.33 & 54.34 & 52.16 & 50.49 & 22.07 \\
    CLOM\cite{zou2022margin} & 73.08 & 68.09 & 64.16 & 60.41 & 57.41 & 54.29 & 51.54 & 49.37 & 48.00 & 25.08 \\
    C-FSCIL\cite{hersche2022constrained} & 76.40 & 71.14 & 66.46 & 63.29 & 60.42 & 57.46 & 54.78 & 53.11 & 51.41 & 24.99\\
    ALICE\cite{peng2022few} & 80.6 & 70.6 & 67.4 & 64.5 & 62.5 & 60.0 & 57.8 & 56.8 & 55.7 & 24.9 \\    
    \hline
    \textbf{BSC} & 81.07 &  76.58 & 72.56 & 69.81 & 67.1 & 64.98 & 63.4 & 61.98 &  \textbf{\underline{60.83}}  &  \textbf{\underline{20.24}} \\
    \thickhline
    \end{tabular}
    \caption{Results of comparative studies on miniImageNet dataset with 5-way 5-shot settings}
    \label{table:sota_miniimagenet}
\end{subtable}}
\vspace*{0.1 mm}
\newline
\resizebox{0.86\textwidth}{!}{%
\begin{subtable}{1\textwidth}
\sisetup{table-format=4.0} 
    \renewcommand{\arraystretch}{1.1}
    \centering
    \begin{tabular}{l c c c c c c c c c  l}
    \thickhline
    \multirow{2}{*}{\textbf{Method}} & \multicolumn{9}{c}{\textbf{Acc. in each session$\uparrow$ (\%) }} & \multirow{2}{*}{\textbf{PD($\downarrow$)}} \\ 
    \cline{2-10}
    & 1 & 2 & 3 & 4 & 5 & 6 & 7 & 8 & 9 \\
    \hline
    TOPIC\cite{tao2020few} & 64.1 & 55.88 & 47.07 & 45.16 & 40.11 & 36.38 & 33.96 & 31.55 & 29.37 & 34.73 \\
    CEC\cite{zhang2021few} & 73.07 & 68.88 & 65.26 & 61.19 & 58.09 & 55.57 & 53.22 & 51.34 & 49.14 & 23.93\\
    LIMIT\cite{zhou2022few} & 73.81 & 72.09 & 67.87 & 63.89 & 60.70 & 57.77 & 55.67 & 53.52 & 51.23 & 22.58 \\
    FACT\cite{zhou2022forward} & 74.60 & 72.09 & 67.56 & 63.52 & 61.38 & 58.36 & 56.28 & 54.24 & 52.10 & 22.50 \\
    CLOM\cite{zou2022margin} & 74.20 & 69.83 & 66.17 & 62.39 & 59.26 & 56.48 & 54.36 & 52.16 & 50.25 & 23.95 \\
    C-FSCIL\cite{hersche2022constrained} & 77.47 & 72.40 & 67.47 & 63.25 & 59.84 & 56.95 & 54.42 & 52.47 & 50.47 & 27.00 \\ 
    ALICE\cite{peng2022few} & 79.0 & 70.5 & 67.1 & 63.4 & 61.2 & 59.2 & 58.1 & 56.3 & 54.1 & 24.9 \\
    \hline
    \textbf{BSC} &  75.88 & 70.29 & 67.93 & 64.5 & 61.55 & 59.98 & 58.28 & 56.38 & \textbf{\underline{55.51}} & \textbf{\underline{20.37}} \\
    \thickhline
    \end{tabular}
    \caption{Results of comparative studies on CIFAR100 dataset with 5-way 5-shot settings}
    \label{table:sota_cifar100}
\end{subtable}}
\newline
\vspace*{0.3 mm}
\newline
\resizebox{0.86\textwidth}{!}{%
\begin{subtable}{1\textwidth}
\sisetup{table-format=-1.2}   
    \renewcommand{\arraystretch}{1.1}
    \centering
    \begin{tabular}{l c c c c c c c c c c c l }
    \thickhline
    \multirow{2}{*}{\textbf{Method}} & \multicolumn{11}{c}{\textbf{Acc. in each session$\uparrow$ (\%) }} & \multirow{2}{*}{\textbf{PD($\downarrow$)}} \\ 
    \cline{2-12}
    & 1 & 2 & 3 & 4 & 5 & 6 & 7 & 8 & 9 & 10 & 11 \\
    \hline
    TOPIC\cite{tao2020few} & 68.68 & 62.49 & 54.81 & 49.99 & 45.25 & 41.4 & 38.35 & 35.36 & 32.22 & 28.31 & 26.28 & 42.40 \\
     CEC\cite{zhang2021few} & 75.85 & 71.94 & 68.50 & 63.5 & 62.43 & 58.27 & 57.73 & 55.81 & 54.83 & 53.52 & 52.28 & 23.57\\
    LIMIT\cite{zhou2022few} & 75.89 & 73.55 & 71.99 & 68.14 & 67.42 & 63.61 & 62.40 & 61.35 & 59.91 & 58.66 & 57.41 & 18.48 \\
    FACT\cite{zhou2022forward} & 75.90 & 73.23 & 70.84 & 66.13 & 65.56 & 62.15 & 61.74 & 59.83 & 58.41 & 57.89 & 56.94 & 18.96 \\
    CLOM\cite{zou2022margin} & 79.57 & 76.07 & 72.94 & 69.82 & 67.80 & 65.56 & 63.94 & 62.59 & 60.62 & 60.34 & 59.58 & 19.99 \\
    ALICE\cite{peng2022few} & 77.4 & 72.7 & 70.6 & 67.2 & 65.9 & 63.4 & 62.9 & 61.9 & 60.5 & 60.6 & 60.1 & 17.3 \\
    \hline
    {\textbf{BSC}} & 80.1 & 76.55 & 73.98 & 71.97 & 70.41 & 70.29 & 69.16 & 66.30 & 65.63 & 64.36 & \textbf{\underline{63.02}} & \textbf{\underline{17.08}} \\
    \thickhline
    \end{tabular}
    \caption{Results of comparative studies on CUB200 dataset using ImageNet-pre-trained model with 10-way 5-shot settings}
    \label{app:table:sota_cub200_pretrain}
\end{subtable}}

\caption{Comparison with the state-of-the-art methods on miniImageNet, CIFAR100, and CUB200 datasets. $\uparrow$ means the higher is the better, while $\downarrow$ denotes the lower is the better. Complete table including additional methods is on our supplementary materials.
}
\label{table:sota}
\end{table*}

\subsection{Experimental Setup}
\label{sec:exp_setups}
\textbf{Dataset.}
We evaluate the compared methods on three well-known datasets, miniImageNet \cite{vinyals2016matching}, CUB200 \cite{wah2011caltech}, and CIFAR100 \cite{krizhevsky2019cifar}.
miniImageNet is the 100-class subset fo the ImageNet-1k \cite{russakovsky2015imagenet} dataset used by few-shot learning. The dataset contains a total of 100 classes, where each class consists of 500 training images and 100 test images. The size of each image is $84\times84$. 
CIFAR100 includes 100 classes, where each class has 600 images, divided into 500 training images and 100 test images. Each image has a size of $32\times32$. 
CUB200 dataset is originally designed for fine-grained image classification introduced by \cite{wah2011caltech} for incremental learning. The dataset contains around 6,000 training images and 6,000 test images over 200 classes of birds. The images are resized to $256\times256$ and then cropped to $224\times224$ for the training.

Following the settings of \cite{tao2020few}, we randomly choose 60 classes as base classes and split the remaining incremental classes into 8 sessions, each with 5 classes for the miniImageNet and CIFAR100.
For CUB200, we split into 100 base classes and 100 incremental classes. The incremental classes are divided into 10 sessions, each with 10 classes. 
For all datasets, 5 training images are drawn for each class as we use 5-shot for the experiments, while we use all test images to evaluate the generalization performance to prevent overfitting. \\

\textbf{Evaluation protocol.}
In the previous works, the primary metric to show the performance of the algorithm is the performance dropping rate (\textbf{PD)}, defined as the absolute test accuracy difference between the base session and the last session.
To be specific, we define $\mathcal{A}^t_\mathcal{C}$ to be the accuracy evaluated only using test sets of classes  $\mathcal{C}$ where probabilities are calculated among $\mathcal{C}^{1:t}$ using the network model trained after session $t$.
Then, PD is defined as $PD = \mathcal{A}^1_{\mathcal{C}^1}-\mathcal{A}^T_{\mathcal{C}^{1:T}}$.
\comment{
\begin{equation}
\begin{split}
PD = \mathcal{A}^1_{\mathcal{C}^1}-\mathcal{A}^T_{\mathcal{C}^{1:T}}.
\end{split}
\label{eq:PD}
\end{equation}
}

It is important to note that PD alone may not be sufficient to demonstrate FSCIL performance. The accuracy $\mathcal{A}_\mathcal{C}$ is the test accuracy averaged over all classes of $\mathcal{C}$. However, since FSCIL has a large number of base classes, only maintaining classification performance on the base sessions can lead to high scores. 
In fact, many previous works exhibit high PD but fail to demonstrate high classification accuracy for new sessions \cite{zhang2021few, tao2020few, cheraghian2021semantic}.
Accordingly, we additionally use the following metrics to measure the ability to learn new tasks (NLA) and maintain the performance of the base task (BMA), respectively:
\begin{equation}
\begin{split}
NLA = \frac{\sum_{t=2}^T\mathcal{A}^t_{\mathcal{C}^{2:t}}}{T-1}, \;\;\;
BMA = \frac{\sum_{t=1}^T\mathcal{A}^t_{\mathcal{C}^{1}}}{T}.
\end{split}
\label{eq:NLABMA}
\end{equation}


\textbf{Implementation details.}
For a fair comparison with the preceding works \cite{zhang2021few, tao2020few}, we employ ResNet18 \cite{he2016deep} for all datasets. We use 3 for augmentation number $m$ in Table \ref{table:sota}. 
Random crop, random scale, color jitter, grayscaling, and random horizontal flip are used for data augmentation candidates during pre-training phase. 
Further details are described in our supplementary materials.

\subsection{Main Results}
\label{sec:main_results}
\noindent\textbf{Comparison with the state-of-the-art methods.}

The performance comparison of the proposed method with state-of-the-art methods is shown in Table \ref{table:sota}. The proposed method outperforms all state-of-the-art methods on all datasets, including miniImageNet, CIFAR100, and CUB200. 
On all datasets, we achieved the highest accuracy on the last session while maintaining the lowest performance dropping rate, with a significant improvement compared to previous works. 
Full table containing the results of additional methods are shown in Appendix \ref{app:sec:add_exps}.




\noindent\textbf{Analysis on the effects of BSC loss on the balancing of generic representations} 

We proceed experiment to analyze the balancing effects, as shown in Table \ref{app:table:bsc_tendency}. 
We investigated the balance between the effectiveness of the extracted representations for the base session and the incremental sessions by observing changes in metrics, BMA and NLA.
For BSC, the results showed the tendency that NLA value to increase and the BMA value to decrease as we raised the $\alpha$ value. This tendency supports our claim that increasing the $\alpha$ value during the feature extractor pre-training may help shift the attention from acquiring information specifically from the given annotations to increasing the generality of the extracted features. We observed that as $\alpha$ increases, NLA increases while BMA decreases. By identifying the optimal $\alpha$ value, we were able to achieve the optimal PD value.

We also compared our approach with other SupCon-based methods. Our results demonstrate that these methods have significantly lower scores compared to BSC, especially for NLA. We believe that the low NLA scores are a result of insufficient generality in the representations used. For PaCo, using additional parametric class-wise learnable centers instead of contrastive loss could potentially reduce the generality of the representations, as shown in Fig. \ref{fig:loss_direction}. In the case of THANOS \cite{chen2022perfectly}, utilizing forced repulsion between \textit{positives-aug} and \textit{positives-diffsrc} may hinder the generalization ability, which is why Grill \textit{et al.} \cite{grill2020bootstrap} presented a self-supervised learning method only with positive pairs for the unlabeled dataset.
Note that $m=2, \alpha=1.0$ case of BSC is identical to the SupCon setting.

\begin{table}[!t]
\renewcommand{\arraystretch}{1.1}
\setlength\tabcolsep{4pt}
\centering
\resizebox{0.90\columnwidth}{!}{%
\begin{tabular}{c | c c| c | c c}
\thickhline
Loss & m & $\alpha$ & PD$(\downarrow)$ & NLA$(\uparrow)$ & BMA$(\uparrow)$\\
\hline
\hline
PaCo* \cite{cui2022generalized} & 2 & - & 28.83 & 27.98 & 67.29 \\
\hline
THANOS* \cite{chen2022perfectly} & 2 & - & 23.76 & 30.55 & 71.58 \\
\hline
\multirow{10}{*}{BSC} & 2 & 1.0 & 23.42  & 32.80 & 71.33 \\
\cline{2-6} 
 & 2 & 1.2 & 22.28  &  34.55 &  71.01 \\
\cline{2-6} 
 & 2 & 1.5 & \textbf{ 21.89} & 35.11 &  71.03 \\
\cline{2-6} 
 & 2 & 2.0 & 23.08 & 35.42 & 70.52 \\
\cline{2-6} 
 & 2 & 4.0 & 28.43 & 40.92 & 61.74 \\
\cline{2-6} 
 & 3 & 1.0 & 21.98 & 35.80 &  70.96 \\
\cline{2-6} 
 & 3 & 1.2 & \textbf{\underline{20.37}}  &  38.46 &  70.83 \\
\cline{2-6} 
 & 3 & 1.5 & 21.29& 36.26 & 70.48 \\
\cline{2-6} 
 & 3 & 2.0 & 22.57 & 38.87 &  69.55 \\
\cline{2-6} 
 & 3 & 4.0 & 23.27 &  41.42 &  68.46 \\
\thickhline
\end{tabular}
}
\caption{Additional results for the analysis on the effects
of BSC loss on the balancing of generic representations.
The experiments were conducted on CIFAR100 dataset.
* indicates the re-implemented results, where we follow the details
from the open-source codes provided.
}
\label{app:table:bsc_tendency}
\end{table}

\noindent\textbf{Analysis on classifier initialization}

Experiments were conducted to demonstrate the importance of consistent initialization methods for both base and incremental classifiers.
We obtain some indicators to understand the phenomenon that occurs when random initialization is used for base classifiers while average feature initialization is used for new classes. Four indicators were analyzed: base session accuracy after fine-tuning, base session accuracy after adding incremental classifiers for the first session, and the ratio of base and new classes test sets, which are classified into one of the incremental classes. Since the feature extractor is not updated after fine-tuning, only newly created classifiers are added for the new session. As shown in Fig. \ref{fig:angles_and_bar}, the ``Base$\rightarrow$New clf ratio" is significantly high. The drop in prediction accuracy is 43.1\% on CIFAR100 and 8.8\% on miniImageNet, both of which are critical to FSCIL performance (the gap between base session accuracy $\mathcal{A}^1_{\mathcal{C}^1}$ and BMA is within 2\% for both datasets). We suppose that the difference in the drop rate between the datasets is caused by the domain distribution. The results show that the features of base classes are comparatively closer to incremental classifiers within the feature space than base classifiers, which are spread all over the embedding space. Note that the feature space is very limited to the narrow area of the embedded space in an angular perspective, as analyzed in Appendix \ref{app:sec:angle_basetrain}. In short, using both random initialization and feature-average initialization causes a severe imbalance between the classifiers.

\noindent\textbf{Visualization of representations for new tasks}

The effectiveness of representations for the incremental classes was visualized using t-SNE analysis, as shown in Figs. \ref{fig:tsne_ce} and \ref{fig:tsne_scl}. It is evident that the representations extracted from the feature extractor trained on contrastive loss exhibit better clarity than those trained on class-wise loss. Note that we used cross-entropy loss for class-wise and BSC loss for contrastive loss.

\subsection{Ablation Studies}
\label{sec:ablation_studies}
The effects of modules utilized in the proposed learning scheme are quantitatively examined via ablation studies.
We compare the results of horizontal lines in Table \ref{table:ablation_modules} to analyze the effect of each module.

\comment{
\begin{table}[!t]
\renewcommand{\arraystretch}{1.1}

\setlength\tabcolsep{4pt}
\centering
\begin{tabular}{c c c| c c c}
\thickhline
Loss & m & $\alpha$ & PD$(\downarrow)$ & NLA$(\uparrow)$ & BMA$(\uparrow)$\\
\hline
\hline
BSC & 3 & 1.0 &  21.98 & 35.80 &  70.96 \\
\hline
BSC & 3 & 1.2 &  20.37  &  38.46 &  70.83 \\
\hline
BSC & 3 & 1.5 &  21.29& 36.26 & 70.48 \\
\hline
BSC & 3 & 2.0 &   22.57 & 38.87 &  69.55 \\
\hline
BSC & 3 & 4.0 &  23.27 &  41.42 &  68.46 \\
\thickhline
\end{tabular}
\caption{Experimental results analyzing the effects of BSC loss on the balance of generic representations are presented. In order to evaluate the validity of the extracted representations for both the base session and the incremental session, which is the primary goal of the generic representation approach, we used the NLA and BMA metrics. The experiments were conducted on the CIFAR100 dataset.}
\label{table:balance_analysis}
\end{table}
}

\begin{figure}
\centering
\begin{subfigure}[b]{.7\linewidth}
\includegraphics[width=\linewidth]{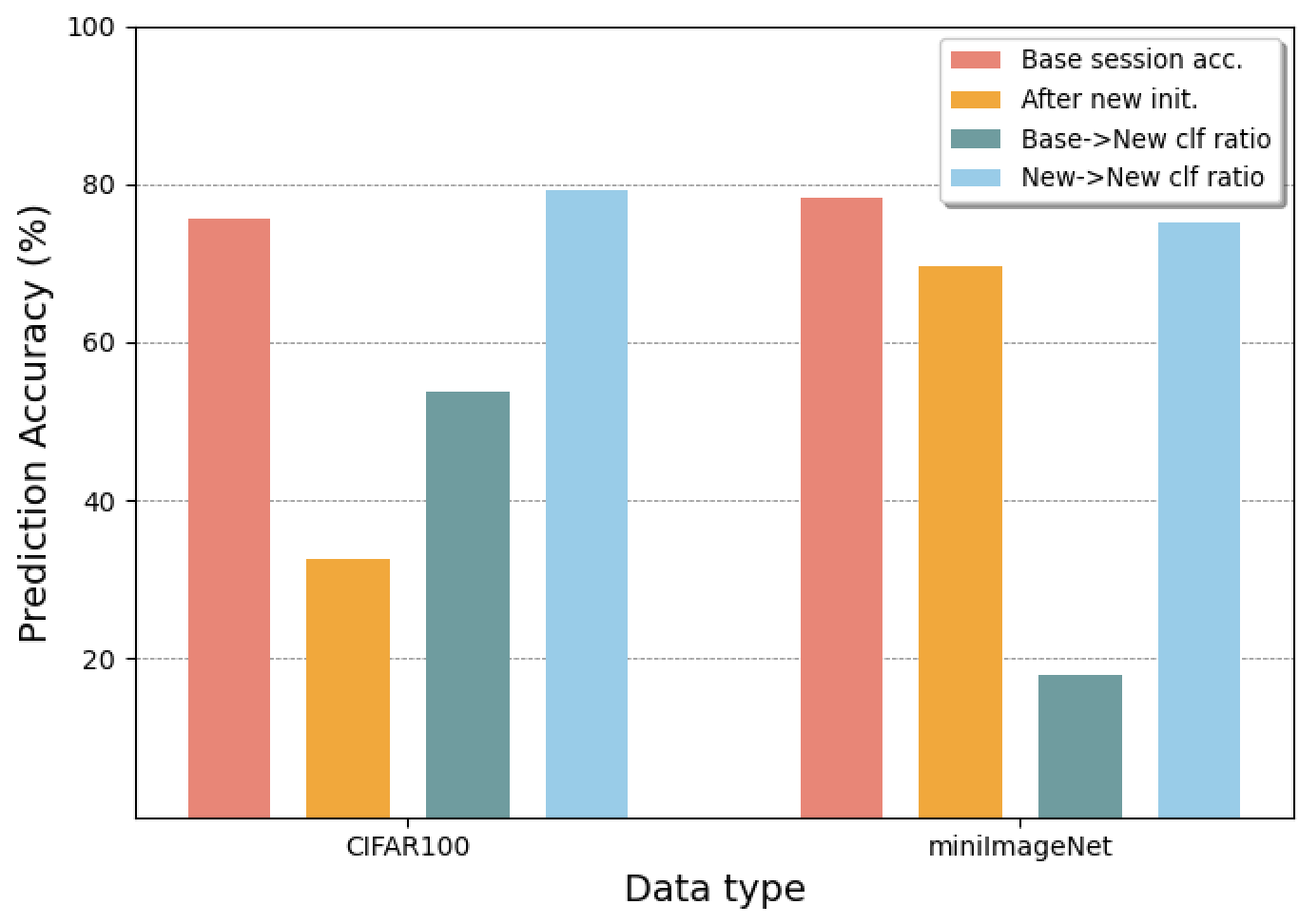}
\centering
\caption{Effects of base classifier random initialization}\label{fig:angles_and_bar}
\end{subfigure}

\begin{subfigure}[b]{.41\linewidth}
\includegraphics[width=\linewidth]{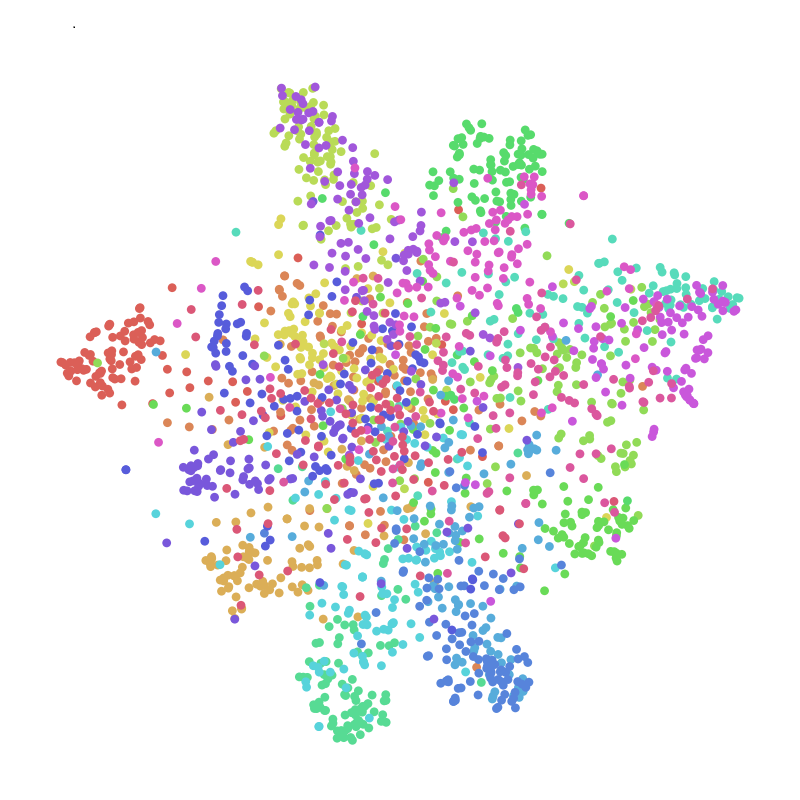}
\caption{Class-wise loss}\label{fig:tsne_ce}
\end{subfigure}
\begin{subfigure}[b]{.41\linewidth}
\includegraphics[width=\linewidth]{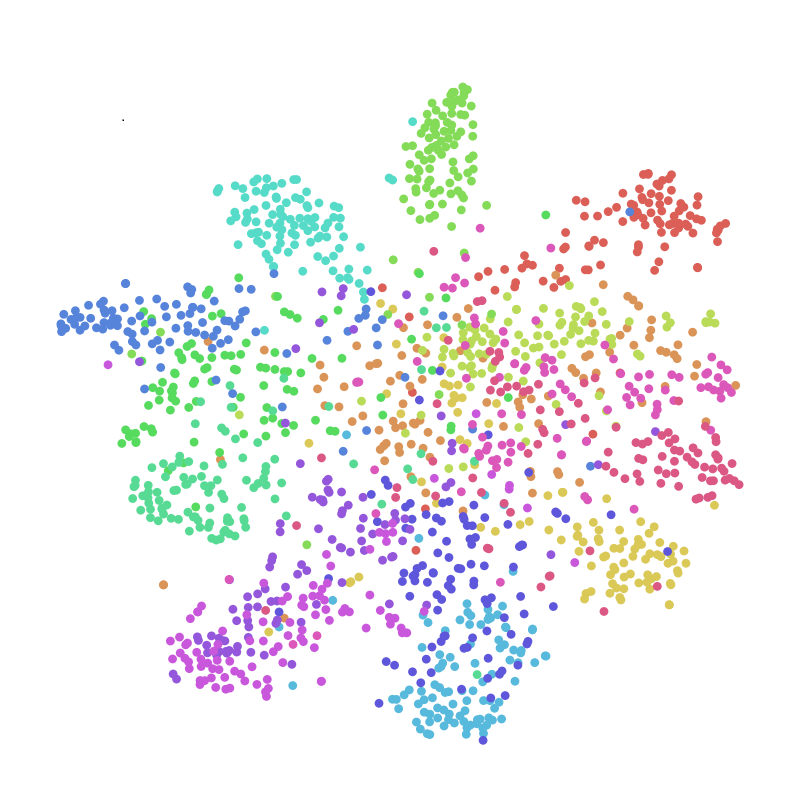}
\caption{Contrastive loss}\label{fig:tsne_scl}
\end{subfigure}
\caption{(a) Intuitive analysis for the effect of adding new classes on the classification performance of base classes. `Base session acc.’ denotes the test accuracy of base session classes after the training of the fine-tuning. `After new init.’ is the classification accuracy measured only on base classes at the first incremental session. `Base$\rightarrow$New clf ratio’ and `New$\rightarrow$New clf ratio’ are the ratio of the base and new session test data classified into classes in the new session. (b),(c) Visualization of the representations from the incremental classes using t-SNE analysis.}
\label{fig:tsne}
\end{figure}

\noindent\textbf{Class-wise loss vs Contrastive loss}: We examined whether using class-wise loss impedes the ability of representations to learn new tasks, as illustrated in Fig. \ref{fig:loss_direction}.
This can be shown from the results of CE loss which shows far low NLA value compared to others.


\noindent\textbf{Pre-training: SimCLR vs BSC}: Next, we discuss the reason for using BSC instead of SimCLR loss.
Representations learned with SimCLR loss showed high NLA but very low BMA, which implies that the knowledge learned for the base session was insufficient. This is because the label information is only used for the small epochs during fine-tuning, without being used for the pre-training stage. The limited usage of label information might have caused such insufficiency, which is the reason for us to utilize BSC for pre-training.

\noindent\textbf{Initializing classifiers as average feature vector}: Result shows that random initialization of the base classifiers before the fine-tuning shows severely low BMA and high NLA. This is because most predictions are concentrated on the new classes, which are initialized as a mean feature vector, also shown in Fig. \ref{fig:angles_and_bar}.

\noindent\textbf{Effect of further modules}: The projection head during the contrastive loss is known to extend the generalization ability of the model. 
The absence of a head projection module can hinder the effects of fine-tuning for the transfer of domain knowledge to the model. 
In fact, NLA, without using a head, showed a far lower score.

We designed the fine-tuning phase to transfer the domain knowledge of the dataset to the model. From the results, we could conclude that adding the fine-tuning phase enhanced the overall PD by sharply increasing the NLA while maintaining the BMA. Such results could verify that fine-tuning process adjusts the model to the specific dataset domain after the pre-training.
Also, using the cs-kd loss showed slight enhancement in both PD and NLA, verifying the efficiacy of reducing the overfitting during the fine-tuning phase.

\begin{table}[!t]
\renewcommand{\arraystretch}{1.1}

\setlength\tabcolsep{4pt}
\centering
\resizebox{0.90\columnwidth}{!}{%
\begin{tabular}{c c c c c | c | c c}
\thickhline
Loss & hd & f.t. & cs & rbc  & PD$(\downarrow)$ & NLA$(\uparrow)$ & BMA$(\uparrow)$\\
\hline
\hline
CE & \xmark & \xmark & \xmark & \cmark & 26.81 & 16.97 & 67.35 \\
\hline
SimCLR & \cmark & \cmark & \cmark & \cmark & 27.03 & 50.54 & 42.73 \\
\hline
BSC & \cmark & \cmark & \cmark& \xmark & 51.69 & 62.54 & 22.83 \\
\hline
BSC & \xmark & \cmark & \cmark& \cmark & 25.42 & 22.13 & 71.24 \\
\hline
BSC & \cmark & \cmark & \xmark & \cmark & 20.97 & 36.54 & 70.76 \\
\hline
BSC & \cmark & \xmark & \xmark & \cmark & 24.74 & 27.14 & 70.18 \\
\hline
BSC & \cmark & \cmark& \cmark & \cmark & \textbf{\underline{20.37}} & 38.46 & 70.83  \\

\thickhline
\end{tabular}
}
\caption{Ablation studies were conducted to analyze the importance of each module used in the proposed framework and to compare the effectiveness of feature extractor training methods in achieving general visual representations. The experiments were conducted on the CIFAR100 dataset, where "hd", "f.t.", "cs", and "rbc" denote whether or not to use a projection head, fine-tuning procedure, cs-kd loss in fine-tuning, and initializing base session classifiers as a feature mean before the fine-tuning.}
\label{table:ablation_modules}
\end{table}


%% file: 6_conclusion.tex
\section{Conclusion}
\label{sec:conclusion}
In this paper, we proposed a few-shot class-incremental learning scheme using generic representations. We composed the learning stages as follows: pre-training for the feature extractor, fine-tuning for domain knowledge transfer, and setting classifiers for the incremental classes. For pre-training, we proposed a balanced supervised contrastive loss to balance the validity of representations between the 
viewable and unviewable classes.
We verified that our method exhibits state-of-the-art performance in preventing performance drops and facilitating new task learning, as demonstrated by our proposed metrics. We also analyzed the necessity of unifying the initialization methods of the classifiers and visualized the effectiveness of our representations on new tasks. Furthermore, we conducted ablation studies to quantitatively demonstrate the effectiveness of each module in our design scheme.

%% file: supp.tex
\part*{Supplementary Material} 
\label{app:sec:supp}

\section{Analysis on angular aspects of features}
\label{app:sec:analysis_features}
We further analyze the angular aspects of features within the embedded space 
to eventually justify the space occupied by the features are actually narrow compared to the entire embedding space 
in angular perspective.
First, we formulate the terms in Section \ref{app:sec:forms}.
Then we provide the inter-class minimum angle aspects in \ref{app:sec:anal_angle_clsnums}.
Finally, we analyze the inter-class angles between the mean features in \ref{app:sec:angle_basetrain}.

\subsection{Formulations}
\label{app:sec:forms}
For a clear explanation, we formulate the terms in this section.
We define the average feature of class $c$ in the base session as:
\begin{equation}
\begin{split}
\bar{z}_{c, \theta} = \frac{1}{|\mathcal{D}_{tr, c}^1|} \sum_{x\in \mathcal{D}_{tr, c}^1}f_\theta(x),
\label{app:eq:feature_angle}
\end{split}
\end{equation}
where $\mathcal{D}_{tr, c}^1=\{x| (x,y)\in\mathcal{D}_{tr}^1 \And y==c \}$, for $c\in\mathcal{C}^1$.
Then, the average of inter-class mean feature angles is defined as:
\begin{equation}
\begin{split}
\psi_\theta = \frac{2}{|\mathcal{C}^1|(|\mathcal{C}^1|-1)}\sum_{1\leq i<j \leq |\mathcal{C}^1|}arccos(\bar{z}_{i,\theta}\cdot\bar{z}_{j,\theta}).
\label{app:eq:avg_feature_angle}
\end{split}
\end{equation}

\subsection{Inter-class minimum angle aspects}
\label{app:sec:anal_angle_clsnums}
\begin{figure}[h]
    \centering 
    \subfloat{\includegraphics[width=5.5cm]{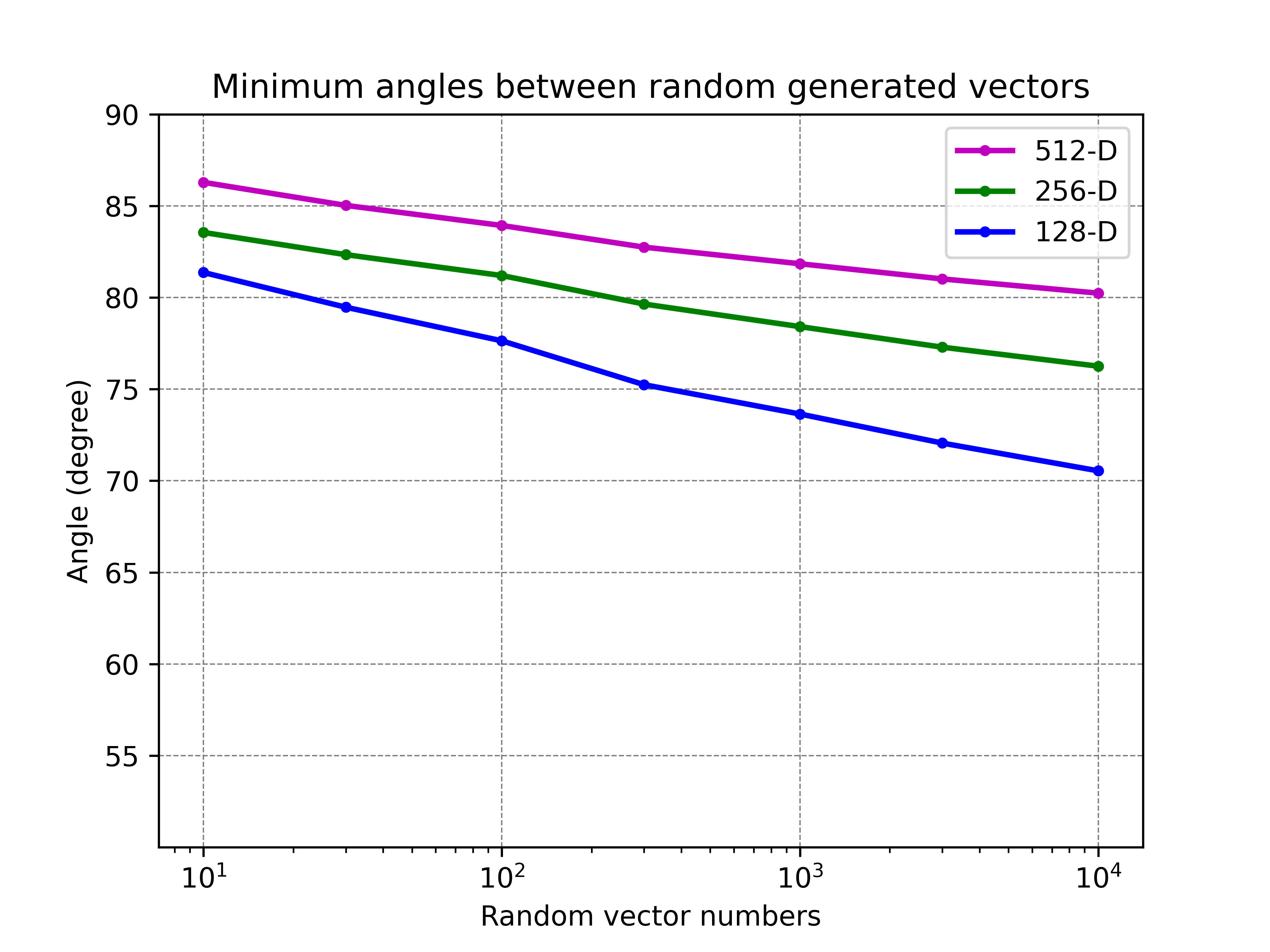}}
    \caption{Minimum angles between the random generated vectors
    }
    \label{app:fig:angle_clsnums}
\end{figure}

To support the statement, we empirically study the minimum angles between the randomly generated vectors.
In details, we randomly generate $n$ vectors $v_i$, $i\in[1,n]$ within the embedded space of dimension $d$. 
Then we measure the average value of minimum angles for each vector, which is defined as: 
\begin{equation}
\begin{split}
\phi(n,d)=\frac{1}{n}\sum_{i=1}^n min_{1\leq j\leq n, i\neq j}arccos(\tilde{v_i}\cdot \tilde{v_j}).
\label{app:eq:min_angle}
\end{split}
\end{equation}

The result of the empirical study is shown in Fig. \ref{app:fig:angle_clsnums}. 
We could verify that more than $10^4$ classes can exist within 512-D embedded space with a minimum angular distance less than 80 degrees ($\leq$80).
Furthermore, according to similar results from Deng \emph{et~al}. \cite{deng2019arcface}, the number of classes could even be increased to $10^8$.
Since we are only using hundreds of classes, we conclude that the angular region occupied by the features is extremely small compared to the entire embedded space after the end of base session training.

\subsection{Inter-class angle aspects between the mean features}
\label{app:sec:angle_basetrain}

\begin{figure}[h]
    \centering 
    \subfloat{\includegraphics[width=5.5cm]{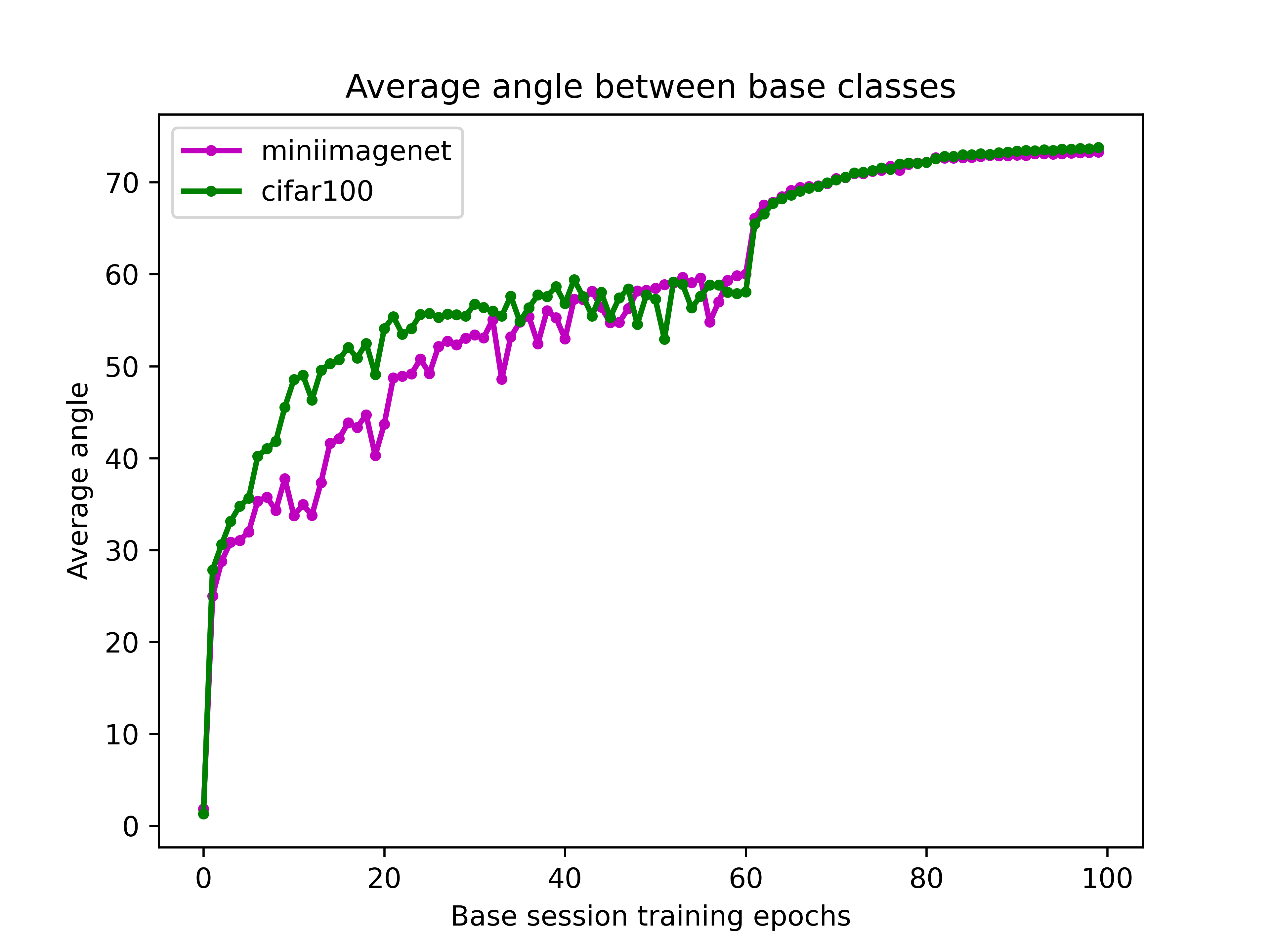}}
    \caption{Average of measured inter-class angles of feature means for each class. The model is randomly initialized at the beginning of the training. The mean feature vectors are gathered at the beginning and dispersed as the training proceeds.
    }
    \label{app:fig:angle_during_training}
\end{figure}

First, we measure the inter-class angles during the feature extractor training using cross-entropy loss.
In Fig. \ref{app:fig:angle_during_training}, the $\psi_\theta$ is nearly zero at the initial epoch of the base session.
This is caused by averaging features with non-zero bias.
In the feature extractor, the final operation passes the input into the ReLU function, which converts all negative terms into zero.
As a result, all features only contain positive elements for each channel. 
In consequence, averaging a large number of features with randomly generated positive-biased values for each channel results in mean feature vectors with similar orientations.
Also, the $\psi_\theta$ is around 73 degrees at the end of the base session.
From the results, we can conclude the angular region covering the features at the end of the base session is significantly small compared to the entire embedded space.

Next, we proceed with the experiment on the BSC case.
When the training feature extractor with BSC loss, we measured the inter-class angular distance after the pre-training and base classifier initialization. The results showed 67 and 79 degrees for the miniImageNet and CIFAR100 datasets. We can come to the same conclusion as in the cross-entropy case with the results.

\begin{table*}[!t]

\sisetup{table-format=-1.2}   
    \renewcommand{\arraystretch}{1.1}
    \centering
    \begin{tabular}{l c c c c c c c c c c c l }
    \thickhline
    \multirow{2}{*}{\textbf{Method}} & \multicolumn{11}{c}{\textbf{Acc. in each session$\uparrow$ (\%) }} & \multirow{2}{*}{\textbf{PD($\downarrow$)}} \\ 
    \cline{2-12}
    & 1 & 2 & 3 & 4 & 5 & 6 & 7 & 8 & 9 & 10 & 11 \\
    \hline
    CEC*\cite{zhang2021few} & 43.23 & 42.12 & 39.24 & 38.49 & 36.78 & 34.86 & 32.75 & 31.67 & 31.12 & 30.41 & 29.2 & 14.03 \\
    FACT*\cite{zhou2022forward} &  31.57 & 27.41 & 24.79 & 22.77 & 22.05 & 20.25 & 19.45 & 18.56 & 17.86 & 17.61 & 16.78 & 14.79 \\
    \hline
    {\textbf{BSC}} &  63.53 & 59.56 & 55.70 & 52.34 & 49.66 & 47.08 & 44.75 & 42.58 & 40.98 & 38.97 & \textbf{\underline{37.30}} & 26.23\\
    \thickhline
    \end{tabular}
\caption{Comparison with state-of-the-art on CUB200 dataset using random initialized model with 10-way 5-shot settings. $\uparrow$ means higher is better, while $\downarrow$ denotes lower is better. * indicates the re-implemented results, where we follow the details from the open-source codes provided.}
\label{app:table:sota_cub200}
\end{table*}

\section{Additional Experiments}
\label{app:sec:cub200}

\subsection{Comparison with state-of-the-art methods}
\label{app:sec:add_exps}
The comparisons of our proposed method with state-of-the-art methods on the CUB200 dataset, using a randomly initialized model, are presented in Table \ref{app:table:sota_cub200}. 
Note that previous works utilized the ImageNet-pre-trained ResNet model for the CUB200 dataset, but using external large-scale datasets might lead to unfair comparisons. Therefore, we present the results without using such pre-trained models in Table \ref{app:table:sota_cub200}. Previous works showed poor results, likely due to the small amount of training data in CUB200 (an average of 30 images per class, whereas MiniImageNet and CIFAR100 have 500 images per class). From the results, it is evident that our proposed method outperforms the previous works significantly.

\begin{table}[!t]
\renewcommand{\arraystretch}{1.0}
\centering
\resizebox{\columnwidth}{!}{%
\begin{tabular}{c | c | c | c c}
\thickhline
Dataset & Method & PD$(\downarrow)$ & NLA$(\uparrow)$ & BMA$(\uparrow)$\\
\hline
\multirow{4}{*}{\shortstack{mini \\ ImageNet}} & CEC* \cite{zhang2021few}& 24.63 & 18.60 & 68.33  \\
& ILDVQ \cite{chen2020incremental}& 22.93  & 13.53 & 62.11 \\
& FACT*\cite{zhou2022forward} & 22.07 & 13.49 & 75.20 \\
& \textbf{BSC} & \textbf{\underline{20.24}}  & \textbf{\underline{37.81}} & \textbf{\underline{77.81}} \\
\hline
\multirow{3}{*}{CIFAR100} & CEC* \cite{zhang2021few}& 23.73 & 23.68 & 67.92 \\
& FACT*\cite{zhou2022forward} & 22.50 & 24.28 & 70.52  \\
& \textbf{BSC} & \textbf{\underline{20.37}}  &  \textbf{\underline{38.46}} & \textbf{\underline{70.83}} \\
\hline
\multirow{4}{*}{CUB200} & CEC* \cite{zhang2021few}& 24.96 & 46.75 & 74.62 \\
& ILDVQ \cite{chen2020incremental}& 19.56 & 42.90 & 75.05 \\
& FACT*\cite{zhou2022forward} & 18.96 & 39.78 & 75.93 \\
& \textbf{BSC} & \textbf{\underline{17.08}} & \textbf{\underline{53.37}} & \textbf{\underline{77.63}} \\
\thickhline
\end{tabular}
}
\caption{Comparison of algorithms with PD, NLA, and BMA metrics. Since NLA and BMA are only reported in Chen \emph{et~al}. \cite{chen2020incremental}, the values for CEC and FACT are re-implemented (*) following the details uploaded on the open-source codes. 
}
\label{table:NLABMA}
\end{table}

\subsection{Comparison on additional metrics}
\label{app:sec:CUBmetrics}

We compared our proposed method with state-of-the-art algorithms, using additional metrics NLA and BMA, to evaluate its classification ability for new session classes and the extent of forgetting on base session classes. The results are summarized in Table \ref{table:NLABMA}. Our proposed method outperformed all other state-of-the-art algorithms across all datasets and metrics. Our high scores on PD and BMA indicate that our method effectively prevents forgetting, while the high NLA score verifies its adaptability to new tasks. Our NLA score was particularly high, demonstrating that our extracted representations work well with new classes. In short, the diverse metrics verify that our extracted representations exhibit the characteristics of generic representations and are effective for all previous, current, and new sessions.



\section{Complete Table of State-of-the-art Methods}
\label{app:sec:fulltable}

We have provided a complete table that compares the performance of previous FSCIL approaches in Table \ref{app:table:sota}. Note that we only included results achieved from a fair comparison setting, which entails using a similar backbone model for feature extraction and not utilizing external datasets.

\begin{table*}[!t]
\centering
\resizebox{0.9\textwidth}{!}{%
\begin{subtable}{1\textwidth}
\sisetup{table-format=4.0} 
    \renewcommand{\arraystretch}{1.1}
    \centering
    \begin{tabular}{l c c c c c c c c c l }
    \thickhline
    \multirow{2}{*}{\textbf{Method}} & \multicolumn{9}{c}{\textbf{Acc. in each session$\uparrow$ (\%) }} & \multirow{2}{*}{\textbf{PD($\downarrow$)}} \\ 
    \cline{2-10}
     & 1 & 2 & 3 & 4 & 5 & 6 & 7 & 8 & 9 \\
    \hline
     TOPIC\cite{tao2020few} & 61.31 & 50.09 & 45.17 & 41.16 & 37.48 & 35.52 & 32.19 & 29.46 & 24.42 & 36.89 \\
     IDLVQ-C\cite{chen2020incremental} &  64.77 & 59.87 & 55.93 & 52.62 & 49.88 & 47.55 & 44.83 & 43.14 & 41.84 & 22.93 \\
    CEC\cite{zhang2021few} & 72.00 & 66.83 & 62.97 & 59.43 & 56.70 & 53.73 & 51.19 & 49.24 & 47.63 & 24.37 \\
    Data replay\cite{liu2022few} & 71.84 & 67.12 & 63.21 & 59.77 & 57.01 & 53.95 & 51.55 & 49.52 & 48.21 & 23.63 \\
    MCNet\cite{ji2022memorizing} & 72.33 & 67.70 & 63.50 & 60.34 & 57.59 & 54.70 & 52.13 & 50.41 & 49.08 & 23.25 \\
    LIMIT\cite{zhou2022few} & 72.32 & 68.47 & 64.30 & 60.78 & 57.95 & 55.07 & 52.70 & 50.72 & 49.19 & 23.13 \\
    FACT\cite{zhou2022forward} & 72.56 & 69.63 & 66.38 & 62.77 & 60.6 & 57.33 & 54.34 & 52.16 & 50.49 & 22.07 \\
    CLOM\cite{zou2022margin} & 73.08 & 68.09 & 64.16 & 60.41 & 57.41 & 54.29 & 51.54 & 49.37 & 48.00 & 25.08 \\
    C-FSCIL\cite{hersche2022constrained} & 76.40 & 71.14 & 66.46 & 63.29 & 60.42 & 57.46 & 54.78 & 53.11 & 51.41 & 24.99\\
    ALICE\cite{peng2022few} & 80.6 & 70.6 & 67.4 & 64.5 & 62.5 & 60.0 & 57.8 & 56.8 & 55.7 & 24.9 \\    
    \hline
    \textbf{BSC (m=2)} & 80.37 &  74.55 & 71.44 & 68.72 & 65.89 & 63.56 & 62.01 & 60.26 & 58.54  &  21.53 \\
    \textbf{BSC (m=3)} & 81.07 &  76.58 & 72.56 & 69.81 & 67.1 & 64.98 & 63.4 & 61.98 &  \textbf{\underline{60.83}}  &  \textbf{\underline{20.24}} \\
    \thickhline
    \end{tabular}
    \caption{Results of comparative studies on miniImageNet dataset with 5-way 5-shot settings}
    \label{table:sota_miniimagenet}
\end{subtable}}
\vspace*{0.1 mm}
\newline
\resizebox{0.9\textwidth}{!}{%
\begin{subtable}{1\textwidth}
\sisetup{table-format=4.0} 
    \renewcommand{\arraystretch}{1.1}
    \centering
    \begin{tabular}{l c c c c c c c c c  l}
    \thickhline
    \multirow{2}{*}{\textbf{Method}} & \multicolumn{9}{c}{\textbf{Acc. in each session$\uparrow$ (\%) }} & \multirow{2}{*}{\textbf{PD($\downarrow$)}} \\ 
    \cline{2-10}
    & 1 & 2 & 3 & 4 & 5 & 6 & 7 & 8 & 9 \\
    \hline
    TOPIC\cite{tao2020few} & 64.1 & 55.88 & 47.07 & 45.16 & 40.11 & 36.38 & 33.96 & 31.55 & 29.37 & 34.73 \\
    ERL++\cite{dong2021few} & 73.62 & 68.22 & 65.14 & 61.84 & 58.35 & 55.54 & 52.51 & 50.16 & 48.23 & 25.39 \\
    CEC\cite{zhang2021few} & 73.07 & 68.88 & 65.26 & 61.19 & 58.09 & 55.57 & 53.22 & 51.34 & 49.14 & 23.93\\
    Data replay\cite{liu2022few} & 74.4 & 70.2 & 66.54 & 62.51 & 59.71 & 56.58 & 54.52 & 52.39 & 50.14 & 24.26\\
    MCNet\cite{ji2022memorizing} & 73.30 & 69.34 & 65.72 & 61.70 & 58.75 & 56.44 & 54.59 & 53.01 & 50.72 & 22.58 \\
    LIMIT\cite{zhou2022few} & 73.81 & 72.09 & 67.87 & 63.89 & 60.70 & 57.77 & 55.67 & 53.52 & 51.23 & 22.58 \\
    FACT\cite{zhou2022forward} & 74.60 & 72.09 & 67.56 & 63.52 & 61.38 & 58.36 & 56.28 & 54.24 & 52.10 & 22.50 \\
    CLOM\cite{zou2022margin} & 74.20 & 69.83 & 66.17 & 62.39 & 59.26 & 56.48 & 54.36 & 52.16 & 50.25 & 23.95 \\
    C-FSCIL\cite{hersche2022constrained} & 77.47 & 72.40 & 67.47 & 63.25 & 59.84 & 56.95 & 54.42 & 52.47 & 50.47 & 27.00 \\ 
    ALICE\cite{peng2022few} & 79.0 & 70.5 & 67.1 & 63.4 & 61.2 & 59.2 & 58.1 & 56.3 & 54.1 & 24.9 \\
    \hline
    \textbf{BSC (m=2)} &  75.53 & 69.79 & 68.34 & 64.74 & 61.96 & 59.75 & 57.4 & 55.24 & 53.64 & 21.89 \\
    \textbf{BSC (m=3)} &  75.88 & 70.29 & 67.93 & 64.5 & 61.55 & 59.98 & 58.28 & 56.38 & \textbf{\underline{55.51}} & \textbf{\underline{20.37}} \\
    \thickhline
    \end{tabular}
    \caption{Results of comparative studies on CIFAR100 dataset with 5-way 5-shot settings}
    \label{table:sota_cifar100}
\end{subtable}}
\newline
\vspace*{0.3 mm}
\newline
\resizebox{0.9\textwidth}{!}{%
\begin{subtable}{1\textwidth}
\sisetup{table-format=-1.2}   
    \renewcommand{\arraystretch}{1.1}
    \centering
    \begin{tabular}{l c c c c c c c c c c c l }
    \thickhline
    \multirow{2}{*}{\textbf{Method}} & \multicolumn{11}{c}{\textbf{Acc. in each session$\uparrow$ (\%) }} & \multirow{2}{*}{\textbf{PD($\downarrow$)}} \\ 
    \cline{2-12}
    & 1 & 2 & 3 & 4 & 5 & 6 & 7 & 8 & 9 & 10 & 11 \\
    \hline
    TOPIC\cite{tao2020few} & 68.68 & 62.49 & 54.81 & 49.99 & 45.25 & 41.4 & 38.35 & 35.36 & 32.22 & 28.31 & 26.28 & 42.40 \\
    IDLVQ-C\cite{chen2020incremental} & 77.37 & 74.72 & 70.28 & 67.13 & 65.34 & 63.52 & 62.10 & 61.54 & 59.04 & 58.68 & 57.81 & 19.56 \\
    SKD\cite{cheraghian2021semantic} & 68.23 & 60.45 & 55.70 & 50.45 & 45.72 & 42.90 & 40.89 & 38.77 & 36.51 & 34.87 & 32.96 & 35.27
     \\
     ERL++\cite{dong2021few} & 73.52 & 71.09 & 66.13 & 63.25 & 59.49 & 59.89 & 58.64 & 57.72 & 56.15 & 54.75 & 52.28 & 21.24 \\
     CEC\cite{zhang2021few} & 75.85 & 71.94 & 68.50 & 63.5 & 62.43 & 58.27 & 57.73 & 55.81 & 54.83 & 53.52 & 52.28 & 23.57\\
    Data replay\cite{liu2022few} & 75.90 & 72.14 & 68.64 & 63.76 & 62.58 & 59.11 & 57.82 & 55.89 & 54.92 & 53.58 & 52.39 & 23.51 \\
    MCNet\cite{ji2022memorizing} &  77.57 & 73.96 & 70.47 & 65.81 & 66.16 & 63.81 & 62.09 & 61.82 & 60.41 & 60.09 & 59.08 & 18.49 \\
    LIMIT\cite{zhou2022few} & 75.89 & 73.55 & 71.99 & 68.14 & 67.42 & 63.61 & 62.40 & 61.35 & 59.91 & 58.66 & 57.41 & 18.48 \\
    FACT\cite{zhou2022forward} & 75.90 & 73.23 & 70.84 & 66.13 & 65.56 & 62.15 & 61.74 & 59.83 & 58.41 & 57.89 & 56.94 & 18.96 \\
    CLOM\cite{zou2022margin} & 79.57 & 76.07 & 72.94 & 69.82 & 67.80 & 65.56 & 63.94 & 62.59 & 60.62 & 60.34 & 59.58 & 19.99 \\
    ALICE\cite{peng2022few} & 77.4 & 72.7 & 70.6 & 67.2 & 65.9 & 63.4 & 62.9 & 61.9 & 60.5 & 60.6 & 60.1 & 17.3 \\
    \hline
    {\textbf{BSC (m=2)}} & 78.37 & 74.64 & 72.02 & 70.49 & 69.54 & 68.33 & 66.24 & 65.57 & 64.90 & 62.81 & 61.03 & 17.34 \\
    {\textbf{BSC (m=3)}} & 80.1 & 76.55 & 73.98 & 71.97 & 70.41 & 70.29 & 69.16 & 66.30 & 65.63 & 64.36 & \textbf{\underline{63.02}} & \textbf{\underline{17.08}} \\
    \thickhline
    \end{tabular}
    \caption{Results of comparative studies on CUB200 dataset using ImageNet-pre-trained model with 10-way 5-shot settings}
    \label{app:table:sota_cub200_pretrain}
\end{subtable}}

\caption{Comparison with the state-of-the-art methods on miniImageNet, CIFAR100, and CUB200 datasets with 5-way 5-shot setting. $\uparrow$ means the higher is the better, while $\downarrow$ denotes the lower is the better. 
}
\label{app:table:sota}
\end{table*}

\section{Further Implementation Details}
\label{app:sec:impl_details}
Within the ResNet18 model, we use a small kernel for conv1 for CIFAR100 and miniImageNet datasets due to the small input size of images \cite{he2016deep}. 
The base model is trained by the SGD optimizer (momentum of 0.9, gamma 0.9 and weight decay of 5e-4). 
The base pre-training epoch is 1000, the initial learning rate is 0.1, and decayed with 0.1 ratio.
In the fine-tuning, we use 10 epochs, learning rate 0.2 with cosine-annealing optimizer.
Classifiers for each class have the same dimensions as the feature extractor output and are initialized with Xavier uniform \cite{glorot2010understanding}. 
The output dimension of the feature extractor is 512. The projection network consists of a 2-layer multi-layered perceptron, projecting the features to a space of 256 dimensions. 
Note that all experiments in our paper represent the average value obtained from 10 independent runs, in order to account for the effects of randomness.

\newpage

\section{Algorithm}
\label{app:sec:alogrithm}

Our overall algorithm is introduced in Algorithm \ref{app:algorithm}.   
\section{Limitations and Future Works}
Our method aims to generate representations with generic characteristics based on a given base session training dataset, which may be biased or contain an insufficient amount of data for learning high-quality generic representations. Limited generalization abilities could lead to performance limitations in more diverse settings, such as incremental sessions with datasets from different data distributions or settings with more incremental classes and sessions. To address such issues, we consider utilizing large-scale pre-trained models for the few-shot class-incremental learning task. This approach could potentially improve generalization by leveraging pre-existing knowledge learned from a more extensive and diverse set of data.   
\bigbreak
\bigbreak
\bigbreak
\bigbreak



\begin{algorithm}
  \caption{Overall learning algorithm}
  
  \begin{algorithmic}[1]
    \Require Image dataset $\mathcal{D}^{1:T}$, random initialized network $f_\theta$, $h_\phi$ and class weights $w^{1:T}$ 
    \Ensure An optimized feature extractor $f_{\theta}$ and class weights $w_{\mathcal{C}^{1:T}}$ for the classification on classes $\mathcal{C}^{1:T}$
    \phase{Base session ($t=1$) training}
    
    \State Initialize $\theta$
    
    \For{Each epoch}
        \State Create multi-viewed batch $J$ from the given training batch, as Eq. \ref{eq:multiview_twotype}.
        \State Calculate loss $\mathcal{L}_{BSC}$ in Eq. \ref{eq:loss_pre}.
        \State Update $\theta, \phi$ and $w_{\mathcal{C}^1}$ with optimizer using loss $\mathcal{L}_{pre}$
    \EndFor
    
    \phase{Base session ($t=1$) fine-tuning}
    Initialize $w_{\mathcal{C}^1}$ following Eq. \ref{eq:base_clf_init}.
    \For{Each epoch}
        \State Create multi-viewed batch $J$ from the given training batch, as Eq. \ref{eq:multiview_twotype}.
        \State Calculate loss $\mathcal{L}_{ft}$ in Eq. \ref{eq:loss_ft}
        \State Update $\theta$ and $w_{\mathcal{C}^1}$ with optimizer using loss $\mathcal{L}_{ft}$
    \EndFor
    \State Test with $\mathcal{D}^1_{te}$ for base session accuracy
    
    \phase{Incremental sessions ($t>1$) training}
        \For{Each session $t$}
            \State Initialize $w_{\mathcal{C}^t}$ following Eq. \ref{eq:inc_clf_init}.
            \State Concatenate $w_{\mathcal{C}^t}$ with $w_{\mathcal{C}^{1:t-1}}$.
            \State Test with $\mathcal{D}_{te}^{1:t}$, $\mathcal{D}_{te}^{1:t-1}$, and $\mathcal{D}_{te}^{2:t}$ for session $t$ accuracy.
        \EndFor
  \end{algorithmic}
  \label{app:algorithm}
\end{algorithm}